# An Exploration of Neural Sequence-to-Sequence Architectures for Automatic Post-Editing

**Marcin Junczys-Dowmunt**
Department for Natural Language Processing
Adam Mickiewicz University in Poznań
junczys@amu.edu.pl

**Roman Grundkiewicz**
School of Informatics
University of Edinburgh
rgrundki@inf.ed.ac.uk

## Abstract

In this work, we explore multiple neural architectures adapted for the task of automatic post-editing of machine translation output. We focus on neural end-to-end models that combine both inputs $mt$ (raw MT output) and $src$ (source language input) in a single neural architecture, modeling $\{mt, src\} \to pe$ directly. Apart from that, we investigate the influence of hard-attention models which seem to be well-suited for monolingual tasks, as well as combinations of both ideas. We report results on data sets provided during the WMT-2016 shared task on automatic post-editing and can demonstrate that dual-attention models that incorporate all available data in the APE scenario in a single model improve on the best shared task system and on all other published results after the shared task. Dual-attention models that are combined with hard attention remain competitive despite applying fewer changes to the input.

## 1 Introduction

Given the raw output of a (possibly unknown) machine translation system from language $src$ to language $mt$, Automatic Post-Editing (APE) is the process of automatic correction of raw MT output ($mt$), so that a closer resemblance to human post-edited MT output ($pe$) is achieved. While APE systems that only model $mt \to pe$ yield good results, the field has always strived towards methods that also integrate $src$ in various forms.

With neural encoder-decoder models, and multi-source models in particular, this can be now achieved in more natural ways than for previously popular phrase-based statistical machine translation (PB-SMT) systems. Despite this, previously reported results for multi-source or dual-source models in APE scenarios are unsatisfying in terms of performance.

In this work, we explore a number of single-source and dual-source neural architectures which we believe to be better fits to the APE task than vanilla encoder-decoder models with soft attention. We focus on neural end-to-end models that combine both inputs $mt$ and $src$ in a single neural architecture, modeling $\{mt, src\} \to pe$ directly. Apart from that, we investigate the influence of hard-attention models, which seem to be well-suited for monolingual tasks. Finally, we create combinations of both architectures.

We report results on data sets provided during the WMT-2016 shared task on automatic post-editing (Bojar et al., 2016) and compare our performance against the shared task winner, the system submitted by the Adam Mickiewicz University (AMU) team (Junczys-Dowmunt and Grundkiewicz, 2016), and a more recent system by Pal et al. (2017) with the previously best published results on the same test set.

Our main contributions are: (1) we perform a thorough comparison of end-to-end neural approaches to APE during which (2) we demonstrate that dual-attention models that incorporate all available data in the APE scenario in a single model achieve the best reported results for the WMT-2016 APE task, and (3) show that models with a hard-attention mechanism reach competitive results although they execute fewer edits than models relying only on soft attention.

The remainder of the paper is organized as follows: Previous relevant work is described in Section 2. Section 3 summarizes the basic encoder-decoder with attention architecture that is further extended with multiple non-standard attention mechanisms in Section 4. These attention mecha-nisms are: hard-attention in Section 4.1, a combination of hard attention and soft attention in Section 4.2, dual soft attention in Section 4.3 and a combination of hard attention and dual soft attention in Section 4.4. We describe experiments and results in Section 5 and conclude in Section 7.

## 2 Previous work

Before the application of neural sequence-to-sequence models to APE, most APE systems would rely on phrase-based SMT following a monolingual approach first introduced by Simard et al. (2007). Béchara et al. (2011) proposed a "source-context aware" variant of this approach where automatically created word alignments were used to create a new source language which consisted of joined MT output and source token pairs. The inclusion of source-language information in that form was shown to improve the automatic post-editing results (Béchara et al., 2012; Chatterjee et al., 2015). The quality of the used word alignments plays an important role for this methods, as demonstrated for instance by Pal et al. (2015).

During the WMT-2016 APE shared task two systems relied on neural models, the CUNI system (Libovický et al., 2016) and the shared task winner, the system submitted by the AMU team (Junczys-Dowmunt and Grundkiewicz, 2016). This submission explored the application of neural translation models to the APE problem and achieved good results by treating different models as components in a log-linear model, allowing for multiple inputs (the source $src$ and the translated sentence $mt$) that were decoded to the same target language (post-edited translation $pe$). Two systems were considered, one using $src$ as the input ($src \rightarrow pe$) and another using $mt$ as the input ($mt \rightarrow pe$). A simple string-matching penalty integrated within the log-linear model was used to control for higher faithfulness with regard to the raw MT output. The penalty fired if the APE system proposed a word in its output that had not been seen in $mt$. The influence of the components on the final result was tuned with Minimum Error Rate Training (Och, 2003) with regard to the task metric TER.

Following the WMT-2016 APE shared task, Pal et al. (2017) published work on another neural APE system that integrated precomputed word-alignment features into the neural structure and enforced symmetric attention during the neural training process. The result was the best reported single neural model for the WMT-2016 APE test set prior to this work. With n-best list re-ranking and combination with phrase-based post-editing systems, the authors improved their results even further. None of their systems, however, integrated information from $src$, all modeled $mt \rightarrow pe$.

## 3 Attentional Encoder-Decoder

Implementations of all models explored in this paper are available in the Marian[1] toolkit (Junczys-Dowmunt et al., 2016). The attentional encoder-decoder model in Marian is a re-implementation of the NMT model in Nematus (Sennrich et al., 2017). The model differs from the standard model introduced by Bahdanau et al. (2015) by several aspects, the most important being the conditional GRU with attention. The summary provided in this section is based on the description in Sennrich et al. (2017).

Given the raw MT output sequence $(x_1, \ldots, x_{T_x})$ of length $T_x$ and its manually post-edited equivalent $(y_1, \ldots, y_{T_y})$ of length $T_y$, we construct the encoder-decoder model using the following formulations.

**Encoder context**   A single forward encoder state $\overrightarrow{\mathbf{h}}_i$ is calculated as:

$$\overrightarrow{\mathbf{h}}_i = \text{GRU}(\overrightarrow{\mathbf{h}}_{i-1}, \mathbf{F}[x_i]),$$

where $\mathbf{F}$ is the encoder embeddings matrix. The GRU RNN cell (Cho et al., 2014) is defined as:

$$\begin{aligned}
\text{GRU}(\mathbf{s}, \mathbf{x}) &= (1 - \mathbf{z}) \odot \underline{\mathbf{s}} + \mathbf{z} \odot \mathbf{s}, \quad (1) \\
\underline{\mathbf{s}} &= \tanh(\mathbf{W}\mathbf{x} + \mathbf{r} \odot \mathbf{U}\mathbf{s}), \\
\mathbf{r} &= \sigma(\mathbf{W}_r\mathbf{x} + \mathbf{U}_r\mathbf{s}), \\
\mathbf{z} &= \sigma(\mathbf{W}_z\mathbf{x} + \mathbf{U}_z\mathbf{s}),
\end{aligned}$$

where $\mathbf{x}$ is the cell input; $\mathbf{s}$ is the previous recurrent state; $\mathbf{W}, \mathbf{U}, \mathbf{W}_r, \mathbf{U}_r, \mathbf{W}_z, \mathbf{U}_z$ are trained model parameters[2]; $\sigma$ is the logistic sigmoid activation function. The backward encoder state is calculated analogously over a reversed input sequence with its own set of trained parameters.

Let $\mathbf{h}_i$ be the annotation of the source symbol at position $i$, obtained by concatenating the forward and backward encoder RNN hidden states, $\mathbf{h}_i = [\overrightarrow{\mathbf{h}}_i; \overleftarrow{\mathbf{h}}_i]$, the set of encoder states $\mathbf{C} = \{\mathbf{h}_1, \ldots, \mathbf{h}_{T_x}\}$ then forms the encoder context.

---
[1] https://github.com/marian-nmt/marian
[2] Biases have been omitted.

**Decoder initialization** The decoder is initialized with start state $s_0$, computed as the average over all encoder states:

$$s_0 = \tanh\left(W_{init} \frac{\sum_{i=1}^{T_x} h_i}{T_x}\right).$$

**Conditional GRU with attention** We follow the Nematus implementation of the conditional GRU with attention, cGRU$_{att}$:

$$s_j = \text{cGRU}_{att}\left(s_{j-1}, E[y_{j-1}], C\right), \quad (2)$$

where $s_j$ is the newly computed hidden state, $s_{j-1}$ is the previous hidden state, C the source context and $E[y_{j-1}]$ is the embedding of the previously decoded symbol $y_{i-1}$.

The conditional GRU cell with attention, cGRU$_{att}$, has a complex internal structure, consisting of three parts: two GRU layers and an intermediate attention mechanism ATT.

Layer GRU$_1$ generates an intermediate representation $s'_j$ from the previous hidden state $s_{j-1}$ and the embedding of the previous decoded symbol $E[y_{j-1}]$:

$$s'_j = \text{GRU}_1\left(s_{j-1}, E[y_{j-1}]\right).$$

The attention mechanism, ATT, inputs the entire context set C along with intermediate hidden state $s'_j$ in order to compute the context vector $c_j$ as follows:

$$c_j = \text{ATT}\left(C, s'_j\right) = \sum_i^{T_x} \alpha_{ij} h_i,$$

$$\alpha_{ij} = \frac{\exp(e_{ij})}{\sum_{k=1}^{T_x} \exp(e_{kj})},$$

$$e_{ij} = v_a^\intercal \tanh\left(U_a s'_j + W_a h_i\right),$$

where $\alpha_{ij}$ is the normalized alignment weight between source symbol at position $i$ and target symbol at position $j$, and $v_a, U_a, W_a$ are trained model parameters.

Layer GRU$_2$ generates $s_j$, the hidden state of the cGRU$_{att}$, from the intermediate representation $s'_j$ and context vector $c_j$:

$$s_j = \text{GRU}_2\left(s'_j, c_j\right).$$

**Deep output** Finally, given $s_j$, $y_{j-1}$, and $c_j$, the output probability $p(y_j|s_j, y_{j-1}, c_j)$ is computed by a softmax activation as follows:

$$p(y_j|s_j, y_{j-1}, c_j) = \text{softmax}\left(t_j W_o\right),$$
$$t_j = \tanh\left(s_j W_{t_1} + E[y_{j-1}] W_{t_2} + c_j W_{t_3}\right).$$

$W_{t_1}, W_{t_2}, W_{t_3}, W_o$ are the trained model parameters.

This rather standard encoder-decoder model with attention is our baseline and denoted as CGRU.

## 4 Encoder-Decoder Models with APE-specific Attention Models

The following models reuse most parts of the architecture described above wherever possible, most differences occur in the decoder RNN cell and the attention mechanism. The encoders are identical, so are the deep output layers.

### 4.1 Hard Monotonic Attention

Aharoni and Goldberg (2016) introduce a simple model for monolingual morphological reinflection with hard monotonic attention. This model looks at one encoder state at a time, starting with the left-most encoder state and progressing to the right until all encoder states have been processed.

The target word vocabulary $V_y$ is extended with a special step symbol ($V'_y = V_y \cup \{\langle\text{STEP}\rangle\}$) and whenever $\langle\text{STEP}\rangle$ is predicted as the output symbol, the hard attention is moved to the next encoder state. Formally, the hard attention mechanism is represented as a precomputed monotonic sequence $(a_1, \ldots, a_{T_y})$ which can be inferred from the target sequence $(y_1, \ldots, y_{T_y})$ (containing original target symbols and $T_x$ step symbols) as follows:

$$a_1 = 1,$$
$$a_j = \begin{cases} a_{j-1} + 1 & \text{if } y_{j-1} = \langle\text{STEP}\rangle \\ a_{j-1} & \text{otherwise.} \end{cases}$$

For a given context $C = \{h_1, \ldots, h_{T_x}\}$, the attended context vector at time step $j$ is simply $h_{a_j}$.

Following the description by Aharoni and Goldberg (2016) for their LSTM-based model, we adapt the previously described encoder-decoder model to incorporate hard attention. Given the sequence of attention indices $(a_1, \ldots, a_{T_y})$, the conditional GRU cell (Eq. 2) used for hidden state updates of the decoder is replaced with a simple GRU cell (Eq. 1) (thus removing the soft-attention mechanism):

$$s_j = \text{GRU}\left(s_{j-1}, \left[E[y_{j-1}]; h_{a_j}\right]\right), \quad (3)$$

where the cell input is now a concatenation of the embedding of the previous target symbol $E[y_{j-1}]$

and the currently attended encoder state $\mathbf{h}_{a_j}$. This model is labeled GRU-HARD.

We find this architecture compelling for monolingual tasks that might require higher faithfulness with regard to the input. With hard monotonic attention, the translation algorithm can enforce certain constraints:

1. The end-of-sentence symbol can only be generated if the hard attention mechanism has reached the end of the input sequence, enforcing full coverage;

2. The ⟨STEP⟩ symbol cannot be generated once the end-of-sentence position in the source has been reached. It is however still possible to generate content tokens.

This model requires a target sequence with correctly inserted ⟨STEP⟩ symbols. For the described APE task, using the Longest Common Subsequence algorithm (Hirschberg, 1977), we first generate a sequence of match, delete and insert operations which transform the raw MT output $(x_1, \cdots x_{T_x})$ into the corrected post-edited sequence $(y_1, \cdots y_{T_y})^3$. Next, we map these operations to the final sequence of steps and target tokens according to the following rules:

- For each matched pair of tokens $x, y$ we produce symbols: ⟨STEP⟩ $y$;

- For each inserted target token $y$ we produce the same token $y$;

- For each deleted source token $x$ we produce ⟨STEP⟩;

- Since at initialization of the model $a_1 = 1$, i.e. the first encoder state is already attended to, we discard the first symbol in the new sequence if it is a ⟨STEP⟩ symbol.

### 4.2 Hard and Soft Attention

While the hard attention model can be used to enforce faithfulness to the original input, we would also like the model to be able to look at information anywhere in the source sequence which is a property of the soft attention model.

By re-introducing the conditional GRU cell with soft attention into the GRU-HARD model while also inputting the hard-attended encoder state $h_{a_j}$, we can try to take advantage of both attention mechanisms. Combining Eq. 2 and Eq. 3, we get:

$$\mathbf{s}_j = \text{cGRU}_{\text{att}}\left(\mathbf{s}_{j-1}, \left[\mathbf{E}[y_{j-1}]; \mathbf{h}_{a_j}\right], \mathbf{C}\right). \quad (4)$$

The rest of the model is unchanged; the translation process is the same as before and we use the same target step/token sequence for training. This model is called CGRU-HARD.

### 4.3 Soft Dual-Attention

Neural multi-source models (Zoph and Knight, 2016) seem to be a natural fit for the APE task as raw MT output and original source language input are available. Although applications to the APE problem have been reported (Libovický and Helcl, 2017), state-of-the-art results seem to be missing.

In this section we give details about our dual-source model implementation. We rename the existing encoder C to $\mathbf{C}^{mt}$ to signal that the first encoder consumes the raw MT output and introduce a structurally identical second encoder $\mathbf{C}^{src} = \{\mathbf{h}_1^{src}, \ldots, \mathbf{h}_{T_{src}}^{src}\}$ over the source language. To compute the decoder start state $s_0$ for the multi-encoder model we concatenate the averaged encoder contexts before mapping them into the decoder state space:

$$\mathbf{s}_0 = \tanh\left(\mathbf{W}_{init}\left[\frac{\sum_{i=1}^{T_{mt}} \mathbf{h}_i^{mt}}{T_{mt}}; \frac{\sum_{i=1}^{T_{src}} \mathbf{h}_i^{src}}{T_{src}}\right]\right).$$

In the decoder, we replace the conditional GRU with attention, with a doubly-attentive cGRU cell (Calixto et al., 2017) over contexts $\mathbf{C}^{mt}$ and $\mathbf{C}^{src}$:

$$\mathbf{s}_j = \text{cGRU}_{\text{2-att}}\left(\mathbf{s}_{j-1}, \mathbf{E}[y_{j-1}], \mathbf{C}^{mt}, \mathbf{C}^{src}\right). \quad (5)$$

The procedure is similar to the original cGRU, differing only in that in order to compute the context vector $\mathbf{c}_j$, we first calculate contexts vectors $\mathbf{c}_j^{mt}$ and $\mathbf{c}_j^{src}$ for each context and then concatenate[4] the results:

---

[3]Similar to GNU `wdiff`.

[4]Calixto et al. (2017) combine their two attention models by modifying their GRU cell to include another set of parameters that is multiplied with the additional context vector and summed in the GRU-components. Formally, both approaches give identical results, as for concatenation the original parameters have to grow in size to match the now longer input vector dimensions. The GRU cell itself does not need to be modified.

$$\mathbf{s}'_j = \text{GRU}_1\left(\mathbf{s}_{j-1}, \mathbf{E}[y_{j-1}]\right),$$

$$\mathbf{c}^{mt}_j = \text{ATT}\left(\mathbf{C}^{mt}, \mathbf{s}'_j\right) = \sum_i^{T_{mt}} \alpha_{ij} \mathbf{h}^{mt}_i,$$

$$\mathbf{c}^{src}_j = \text{ATT}\left(\mathbf{C}^{src}, \mathbf{s}'_j\right) = \sum_i^{T_{src}} \alpha_{ij} \mathbf{h}^{src}_i,$$

$$\mathbf{c}_j = \left[\mathbf{c}^{mt}_j; \mathbf{c}^{src}_j\right],$$

$$\mathbf{s}_j = \text{GRU}_2\left(\mathbf{s}'_j, \mathbf{c}_j\right).$$

This could be easily extended to an arbitrary number of encoders with different architectures. During training, this model is fed with a tri-parallel corpus, and during translation both input sequences are processed simultaneously to produce the corrected output. This model is denoted as M-CGRU.

### 4.4 Hard Attention with Soft Dual-Attention

Analogously to the procedure described in section 4.2, we can extend the doubly-attentive cGRU to take the hard-attended encoder context as additional input:

$$\mathbf{s}_j = \text{cGRU}_{\text{2-att}}\left(\mathbf{s}_{j-1}, \left[\mathbf{E}[y_{j-1}]; \mathbf{h}^{mt}_{a_j}\right], \mathbf{C}^{mt}, \mathbf{C}^{src}\right).$$

In this formulation, only the first encoder context $\mathbf{C}^{mt}$ is attended to by the hard monotonic attention mechanism. The target training data consists of the step/token sequences used for all previous hard-attention models. We call this model M-CGRU-HARD.

## 5 Experiments and Results

### 5.1 Training, Development, and Test Data

We perform all our experiments[5] with the official WMT-2016 (Bojar et al., 2016) automatic post-editing data and the respective development and test sets. The training data consists of a small set of 12,000 post-editing triplets $(src, mt, pe)$, where $src$ is the original English text, $mt$ is the raw MT output generated by an English-to-German system, and $pe$ is the human post-edited MT output. The MT system used to produce the raw MT output is unknown, so is the original training data. The task consists of automatically correcting the MT output so that it resembles human post-edited data. The main task metric is TER (Snover et al., 2006) — the lower the better — with BLEU (Papineni et al., 2002) as a secondary metric.

---

[5] All experiments in this sections can be reproduced following the instructions on https://marian-nmt.github.io/examples/exploration/.

| Data set | Sentences | TER |
|---|---:|---:|
| training set | 12,000 | 26.22 |
| development set | 1,000 | 24.81 |
| test set | 2,000 | – |
| artificial-large | 4,335,715 | 36.63 |
| artificial-small | 531,839 | 25.28 |

Table 1: Statistics for artificial data sets in comparison to official training and development data. Adapted from Junczys-Dowmunt and Grundkiewicz (2016).

To overcome the problem of too little training data, Junczys-Dowmunt and Grundkiewicz (2016) — the authors of the best WMT-2016 APE shared task system — generated large amounts of artificial data via round-trip translations. The artificial data has been filtered to match the HTER statistics of the training and development data for the shared task and was made available for download[6]. Table 1 summarizes the data sets used in this work.

To produce our final training data set we over-sample the original training data 20 times and add both artificial data sets. This results in a total of slightly more than 5M training triplets. We validate on the development set for early stopping and report results on the WMT-2016 test set. The data is already tokenized. Additionally we true-case all files and apply segmentation into BPE subword units (Sennrich et al., 2016). We reuse the subword units distributed with the artificial data set. For the hard-attention models, we create target training and development files following the LCS-based procedure outlined in section 4.1.

### 5.2 Training parameters

All models are trained on the same training data. Models with single input encoders take only the raw MT output ($mt$) as input, dual-encoder models use raw MT output ($mt$) and the original source ($pe$). The training procedures and model settings are the same whenever possible:

---

[6] The artificial filtered data has been made available at https://github.com/emjotde/amunmt/wiki/AmuNMT-for-Automatic-Post-Editing.

|  | dev 2016 | | test 2016 | |
| --- | --- | --- | --- | --- |
| Model | TER↓ | BLEU↑ | TER↓ | BLEU↑ |
| WMT-2016 BASELINE-1 (Bojar et al., 2016) | 25.14 | 62.92 | 24.76 | 62.11 |
| WMT-2016 BASELINE-2 (Bojar et al., 2016) | – | – | 24.64 | 63.47 |
| Junczys-Dowmunt and Grundkiewicz (2016) | 21.46 | 68.94 | 21.52 | 67.65 |
| Pal et al. (2017) SYMMETRIC | – | – | 21.07 | 67.87 |
| Pal et al. (2017) RERANKING | – | – | 20.70 | **69.90** |

Table 2: Results from the literature for the WMT-2016 APE development and test set.

|  | dev 2016 | | test 2016 | |
| --- | --- | --- | --- | --- |
| Model | TER↓ | BLEU↑ | TER↓ | BLEU↑ |
| CGRU | 22.01 | 68.11 | 22.27 | 66.90 |
| GRU-HARD | 22.72 | 66.82 | 22.72 | 65.86 |
| CGRU-HARD | 22.11 | 67.82 | 22.10 | 67.15 |
| M-CGRU | 20.79 | 69.28 | 20.69 | 68.56 |
| M-CGRU × 4 | 20.10 | **70.24** | **19.92** | 69.40 |
| M-CGRU-HARD | 20.83 | 69.02 | 20.87 | 68.14 |
| M-CGRU-HARD × 4 | **20.08** | 70.05 | 20.34 | 68.96 |

Table 3: Results for models explored in this work. Models with × 4 are ensembles of four models. The main WMT 2016 APE shared task metric was TER (the lower the better).

- All embedding vectors consist of 512 units; the RNN states use 1024 units. We choose a vocabulary size of 40,000 for all inputs and outputs. When hard attention models are trained the maximum sentence length is 100 to accommodate the additional step symbols, otherwise 50.

- To avoid overfitting, we use pervasive dropout (Gal and Ghahramani, 2016) over GRU steps and input embeddings, with dropout probabilities 0.2, and over source and target words with probabilities 0.2.

- We use Adam (Kingma and Ba, 2014) as our optimizer, with a mini-batch size of 64. All models are trained with Asynchronous SGD (Adam) on three to four GPUs.

- We train all models until convergence (early-stopping with a patience of 10 based on development set cross-entropy cost), saving model checkpoints every 10,000 mini-batches. For different models we observed early stopping to be triggered between 600,000 and 900,000 mini-batch updates or between 8 and 11 epochs.

- The best eight model checkpoints w.r.t. development set cross-entropy of each training run are averaged element-wise (Junczys-Dowmunt et al., 2016) resulting in new single models with generally improved performance.

- For the multi-source models we repeat the mentioned procedure four times with different randomly initialized weights.

Training time for one model on four NVIDIA GTX 1080 GPUs or NVIDIA TITAN X (Pascal) GPUs is between one and two days, depending on model complexity. The M-CGRU-HARD model is the most complex and trains longest.

### 5.3 Evaluation

Table 2 contains relevant results for the WMT-2016 APE shared task — during the task and afterwards. WMT-2016 BASELINE-1 is the raw uncorrected MT output. BASELINE-2 is the result of a vanilla phrase-based Moses system (Koehn et al., 2007) trained only on the official 12,000 sentences. Junczys-Dowmunt and Grundkiewicz (2016) is the best system at the shared task. Pal et al. (2017) SYMMETRIC is the currently best reported result on the WMT-2016 APE test set for a single neural model (single source), whereas Pal et al. (2017) RERANKING — the overall best reported result on the test set — is a system combination of Pal et al. (2017) SYMMETRIC with phrase-based models via n-best list re-ranking.

| Model | TER-$pe$ | TER-$mt$ |
|---|---|---|
| CGRU | 22.27 | 12.01 |
| GRU-HARD | 22.72 | 9.48 |
| CGRU-HARD | 22.10 | 11.57 |
| M-CGRU | 20.69 | 15.98 |
| M-CGRU $\times$ 4 | 19.92 | 15.41 |
| M-CGRU-HARD | 20.87 | 13.62 |
| M-CGRU-HARD $\times$ 4 | 20.34 | 13.34 |

Table 4: TER w.r.t. the reference compared to TER w.r.t. the input on test 2016. Lower results for TER-$mt$ indicate greater similarity to the input.

| Model | Mod. | Imp. | Det. |
|---|---|---|---|
| CGRU | 1575 | 871 | 399 |
| GRU-HARD | 1479 | 783 | 362 |
| CGRU-HARD | 1564 | 897 | 371 |
| M-CGRU | 1668 | 1020 | 379 |
| M-CGRU $\times$ 4 | 1612 | 1037 | 322 |
| M-CGRU-HARD | 1688 | 1044 | 388 |
| M-CGRU-HARD $\times$ 4 | 1672 | 1074 | 341 |

Table 5: Number of test set sentences modified, improved and deteriorated by each model.

In Table 3 we present the results for the models discussed in this work. Unsurprisingly, none of the single attention models can compete with the better systems reported in the literature. The encoder-decoder model with only hard monotonic attention (GRU-HARD) is the clear loser, while the comparison between CGRU and CGRU-HARD remains inconclusive. CGRU-HARD seems to generalize slightly better, but would not have been chosen based on the development set performance.

The dual-attention models each outperform the best WMT-2016 system and the currently reported best single-model Pal et al. (2017) SYMMETRIC. The ensembles also beat the system combination Pal et al. (2017) RERANKING in terms of TER (not in terms of BLEU though). The simpler dual-attention model with no hard-attention M-CGRU reaches slightly better results on the test set than its counterpart with added hard attention M-CGRU-HARD, but the situation would have been less clear if only the development set were used to determine the best model. The hard-attention model with dual soft-attention benefits less from ensembling.

# 6 Analysis

## 6.1 Faithfulness and Errors

We postulated that the hard-attention models might have a potential for higher faithfulness. Since the APE task is a mostly monolingual task, we can verify this by comparing TER scores with regard to the reference post-edition (TER-$pe$) and TER scores with regard to the raw MT output (TER-$mt$). The lower the TER-$mt$ score the fewer changes have been made to the input to arrive at the output, thus resulting in higher faithfulness. Table 4 contains this comparison for the WMT-2016 APE test set. The hard-attention models indeed make fewer changes than their soft-attention counterparts. This difference is especially dramatic for M-CGRU and M-CGRU-HARD, where only small differences in TER-$pe$ occur, but a gap of more than two TER points for TER-$mt$. This shows that hard-attention models can reach similar TER scores to soft-attention models while performing fewer changes. It might also explain why ensembling has a lower impact on the hard-attention models: higher faithfulness means less variety which results in smaller benefits from ensembles.

Table 5 compares the number of modified, improved and deteriorated test set sentences (2000 in total) for all models. The majority of sentences is being modified. While the number of deteriorated sentences is comparable between models, the number of improved sentences increases for the dual-source architectures. Ensembles lower the number of deteriorated sentences rather than increasing the number of improved sentences.

## 6.2 Visualization of Attention Types

Figures 1 and 2 visualize the behavior of the presented attention variants examined in this work for the example sentences in Table 6.

For this sentence, the unseen MT system mistranslated the word "Set" as "festlegen". The monolingual $mt \rightarrow pe$ systems cannot easily correct the error as the original meaning is lost, but they improve grammaticality. In Figure 1, we see how the soft attention model (CGRU) follows the input roughly monotonically. The monotonic hard attention model (GRU-HARD) does this naturally. For CGRU-HARD, it is interesting to see how the monotonic attention now allows the soft attention mechanism to look around the input sentence more freely or to remain inactive instead of following the monotonic path.

| | |
|---|---|
| *mt* | Wählen Sie einen Tastaturbefehlssatz im Menü <span style="color:red">festlegen</span> . |
| *src* | Select a shortcut set in the <span style="color:blue">Set</span> menu . |
| CGRU | Wählen Sie einen Tastaturbefehlssatz im Menü aus . |
| GRU-HARD | Wählen Sie einen Tastaturbefehlssatz im Menü aus . |
| CGRU-HARD | Wählen Sie einen Tastaturbefehlssatz im Menü aus . |
| M-CGRU | Wählen Sie einen Tastaturbefehlssatz im Menü " <span style="color:green">Satz</span> " aus . |
| M-CGRU-HARD | Wählen Sie einen Tastaturbefehlssatz im Menü " <span style="color:green">Satz .</span> " |
| *pe* | Wählen Sie einen Tastaturbefehlssatz im Menü " <span style="color:green">Satz .</span> " |

Table 6: Example corrections for different models. Only the multi-source models manage to restore the missing translation for "Set" and insert quotes. The added particle "aus" does not appear in the reference, but is grammatically correct as well.

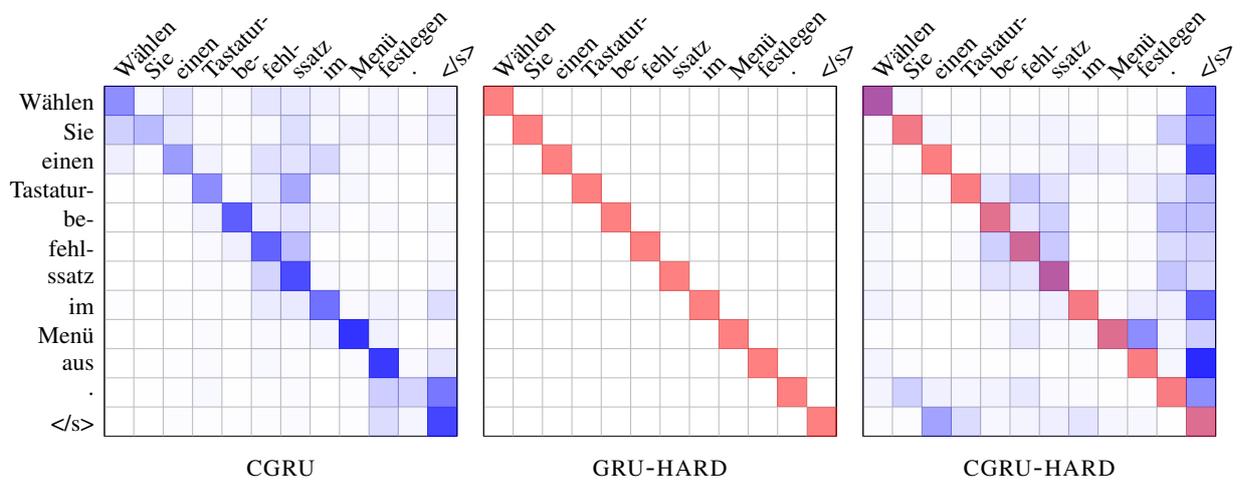

Figure 1: Behavior of different monolingual attention models (best viewed in color).

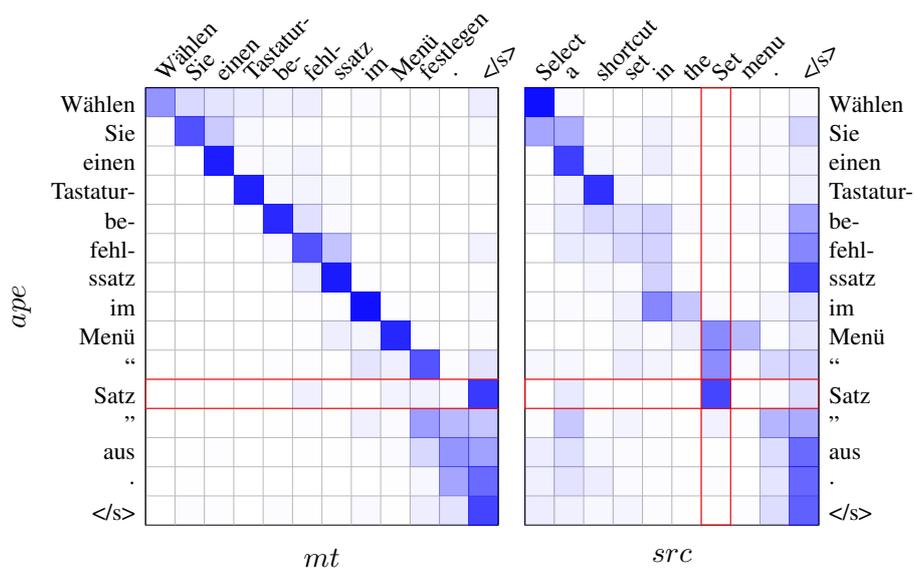

Figure 2: Attention matrices for dual-soft-attention model M-CGRU (best viewed in color).

Both $\{mt, src\} \to pe$ systems take advantage of the $src$ information and improve the input. The proposed modifications could be accepted as correct; one matches the reference. The highlighted rows and columns in Figure 2 show how the original source was used to reconstruct the missing word "Satz" and how both attention mechanisms interact. The attention over $src$ seems to spend most time in a "parking" position at the sentence end unless it can provide useful information; the attention over $mt$ follows the input closely.

## 7 Conclusions and Future Work

In this paper we presented several neural APE models that are equipped with non-standard attention mechanisms and combinations thereof. Among these, hard attention models have been applied to APE for the first time, whereas dual-soft attention models have been proposed before for APE tasks, but with non-conclusive results.

This is the first work to report state-of-the-art results for dual-attention models that integrate full post-edition triplets into a single end-to-end model. The ensembles of dual-attention models provide more than 1.52 TER points improvement over the best WMT-2016 system and 0.7 TER improvement over the best reported system combination for the same test set.

We also demonstrated that while hard-attention models yield similar results to pure soft-attention models, they do so by performing fewer changes to the input. This might be a useful property in scenarios where conservative edits are preferred.

## Acknowledgments

This research was funded by the Amazon Academic Research Awards program. This project has received funding from the European Union's Horizon 2020 research and innovation program under grant 644333 (TraMOOC) and 645487 (ModernMT). This work was partially funded by Facebook. The views and conclusions contained herein are those of the authors and should not be interpreted as necessarily representing the official policies or endorsements, either expressed or implied, of Facebook.

## References

Roee Aharoni and Yoav Goldberg. 2016. Sequence to sequence transduction with hard monotonic attention. *arXiv preprint arXiv:1611.01487* https://arxiv.org/pdf/1611.01487.pdf.

Dzmitry Bahdanau, Kyunghyun Cho, and Yoshua Bengio. 2015. Neural machine translation by jointly learning to align and translate. In *Proceedings of the International Conference on Learning Representations*. San Diego, CA. https://arxiv.org/pdf/1409.0473.pdf.

Hanna Béchara, Yanjun Ma, and Josef van Genabith. 2011. Statistical post-editing for a statistical MT system. In *Proceedings of the 13th Machine Translation Summit*. Xiamen, China, pages 308–315.

Hanna Béchara, Raphaël Rubino, Yifan He, Yanjun Ma, and Josef van Genabith. 2012. An evaluation of statistical post-editing systems applied to RBMT and SMT systems. In *Proceedings of COLING 2012*. Mumbai, India, pages 215–230. http://www.aclweb.org/anthology/C12-1014.

Ondrej Bojar, Rajen Chatterjee, Christian Federmann, Yvette Graham, Barry Haddow, Matthias Huck, Antonio Jimeno Yepes, Philipp Koehn, Varvara Logacheva, Christof Monz, Matteo Negri, Aurelie Neveol, Mariana Neves, Martin Popel, Matt Post, Raphael Rubino, Carolina Scarton, Lucia Specia, Marco Turchi, Karin Verspoor, and Marcos Zampieri. 2016. Findings of the 2016 conference on machine translation. In *Proceedings of the First Conference on Machine Translation*. Association for Computational Linguistics, Berlin, Germany, pages 131–198. http://www.aclweb.org/anthology/W/W16/W16-2301.

Iacer Calixto, Qun Liu, and Nick Campbell. 2017. Doubly-attentive decoder for multi-modal neural machine translation. *CoRR* abs/1702.01287. http://arxiv.org/abs/1702.01287.

Rajen Chatterjee, Marion Weller, Matteo Negri, and Marco Turchi. 2015. Exploring the planet of the APEs: a comparative study of state-of-the-art methods for MT automatic post-editing. In *Proceedings of the 53rd Annual Meeting of the Association for Computational Linguistics and the 7th International Joint Conference on Natural Language Processing*. Association for Computational Linguistics, Beijing, China, pages 156–161. http://www.aclweb.org/anthology/P15-2026.

Kyunghyun Cho, Bart Van Merriënboer, Caglar Gulcehre, Dzmitry Bahdanau, Fethi Bougares, Holger Schwenk, and Yoshua Bengio. 2014. Learning Phrase Representations Using RNN Encoder-Decoder for Statistical Machine Translation. In *Proc. of Empirical Methods in Natural Language Processing*.

Yarin Gal and Zoubin Ghahramani. 2016. A theoretically grounded application of dropout in recurrent neural networks. In *Advances in Neural Information Processing Systems 29 (NIPS)*. https://arxiv.org/pdf/1512.05287.pdf.

Daniel S. Hirschberg. 1977. Algorithms for the longest common subsequence problem. *J. ACM* 24(4):664–675.

Marcin Junczys-Dowmunt, Tomasz Dwojak, and Hieu Hoang. 2016. Is neural machine translation ready for deployment? A case study on 30 translation directions. In *Proceedings of the 9th International Workshop on Spoken Language Translation (IWSLT)*. Seattle, WA.

Marcin Junczys-Dowmunt and Roman Grundkiewicz. 2016. Log-linear combinations of monolingual and bilingual neural machine translation models for automatic post-editing. In *Proceedings of the First Conference on Machine Translation*. pages 751–758. http://www.aclweb.org/anthology/W16-2378.

Diederik P. Kingma and Jimmy Ba. 2014. Adam: A method for stochastic optimization. In *Proceedings of the 3rd International Conference on Learning Representations (ICLR)*. http://arxiv.org/abs/1412.6980.

Philipp Koehn, Hieu Hoang, Alexandra Birch, Chris Callison-Burch, Marcello Federico, Nicola Bertoldi, Brooke Cowan, Wade Shen, Christine Moran, Richard Zens, et al. 2007. Moses: Open source toolkit for statistical machine translation. In *Proceedings of the 45th Annual Meeting of the Association for Computational Linguistics*. Association for Computational Linguistics, pages 177–180.

Jindrich Libovický and Jindrich Helcl. 2017. Attention strategies for multi-source sequence-to-sequence learning. *CoRR* abs/1704.06567. http://arxiv.org/abs/1704.06567.

Jindrich Libovický, Jindrich Helcl, Marek Tlustý, Ondrej Bojar, and Pavel Pecina. 2016. CUNI system for WMT16 automatic post-editing and multimodal translation tasks. In *Proceedings of the First Conference on Machine Translation*. Association for Computational Linguistics, Berlin, Germany, pages 646–654. http://www.aclweb.org/anthology/W/W16/W16-2361.

Franz Josef Och. 2003. Minimum error rate training in statistical machine translation. In *Proceedings of the 41st Annual Meeting of the Association for Computational Linguistics*. Association for Computational Linguistics, Sapporo, Japan, pages 160–167. http://www.aclweb.org/anthology/P03-1021.

Santanu Pal, Sudip Kumar Naskar, Mihaela Vela, Qun Liu, and Josef van Genabith. 2017. Neural automatic post-editing using prior alignment and reranking. In *Proceedings of the European Chapter of the Association for Computational Linguistics*. pages 349–355.

Santanu Pal, Mihaela Vela, Sudip Kumar Naskar, and Josef van Genabith. 2015. USAAR-SAPE: An English–Spanish statistical automatic post-editing system. In *Proceedings of the Tenth Workshop on Statistical Machine Translation*. Association for Computational Linguistics, Lisbon, Portugal, pages 216–221. http://aclweb.org/anthology/W15-3026.

Kishore Papineni, Salim Roukos, Todd Ward, and Wei-Jing Zhu. 2002. BLEU: A method for automatic evaluation of machine translation. In *Proceedings of the 40th Annual Meeting on Association for Computational Linguistics*. Association for Computational Linguistics, Stroudsburg, PA, USA, ACL '02, pages 311–318.

Rico Sennrich, Orhan Firat, Kyunghyun Cho, Alexandra Birch, Barry Haddow, Julian Hitschler, Marcin Junczys-Dowmunt, Samuel Läubli, Antonio Valerio Miceli Barone, Jozef Mokry, and Maria Nadejde. 2017. Nematus: a toolkit for neural machine translation. In *Proceedings of the Software Demonstrations of the 15th Conference of the European Chapter of the Association for Computational Linguistics*. Association for Computational Linguistics, Valencia, Spain, pages 65–68. http://aclweb.org/anthology/E17-3017.

Rico Sennrich, Barry Haddow, and Alexandra Birch. 2016. Neural machine translation of rare words with subword units. In *Proceedings of the 54th Annual Meeting of the Association for Computational Linguistics*. Association for Computational Linguistics, Berlin, Germany, pages 1715–1725.

Michel Simard, Cyril Goutte, and Pierre Isabelle. 2007. Statistical phrase-based post-editing. In *Proceedings of the Conference of the North American Chapter of the Association for Computational Linguistics*. Association for Computational Linguistics, Rochester, New York, pages 508–515.

Matthew Snover, Bonnie Dorr, Richard Schwartz, Linnea Micciulla, and John Makhoul. 2006. A study of translation edit rate with targeted human annotation. In *Proceedings of Association for Machine Translation in the Americas*.

Barret Zoph and Kevin Knight. 2016. Multi-source neural translation. In *Proceedings of the 2016 Conference of the North American Chapter of the Association for Computational Linguistics: Human Language Technologies*. Association for Computational Linguistics, San Diego, California, pages 30–34. http://www.aclweb.org/anthology/N16-1004.